\documentclass[journal]{IEEEtran}
\usepackage{amsmath,amsfonts}
\usepackage{algorithmic}
\usepackage{algorithm}
\usepackage{array}
\usepackage[caption=false,font=normalsize,labelfont=sf,textfont=sf]{subfig}
\usepackage{textcomp}
\usepackage{stfloats}
\usepackage{url}
\usepackage{verbatim}
\usepackage{graphicx}
\usepackage{cite}
\usepackage{booktabs}
\usepackage{multirow}
\usepackage{tabularx}
\usepackage{dsfont}
\usepackage{amssymb}
\usepackage{diagbox}
\usepackage{color}
\usepackage{soul}
\usepackage{pifont}
\usepackage{enumerate}
\usepackage{hyperref}
\usepackage{ tipa }
\usepackage{ wasysym }
\hypersetup{hidelinks,
	colorlinks=true,
	allcolors=black,
	pdfstartview=Fit,
	breaklinks=true}

\begin{document}

\title{Self-supervised Gait-based Emotion Representation Learning from Selective Strongly Augmented Skeleton Sequences}

\author{Cheng~Song,
        Lu~Lu,
        Zhen~Ke,
        Long~Gao,
        Shuai~Ding
\thanks{C.Song, L.Lu, Z.Ke, L.Gao and S.Ding are with the School of Management, Hefei University of Technology. (e-mail: songcheng@hfut.edu.cn; luluccc0317@gmail.com; 2020110762@mail.hfut.edu.cn; 2023111033@mail.hfut.edu.cn; dingshuai@hfut.edu.cn).}}

\maketitle

\begin{abstract}
Emotion recognition is an important part of affective computing. Extracting emotional cues from human gaits yields benefits such as natural interaction, a nonintrusive nature, and remote detection. Recently, the introduction of self-supervised learning techniques offers a practical solution to the issues arising from the scarcity of labeled data in the field of gait-based emotion recognition. However, due to the limited diversity of gaits and the incompleteness of feature representations for skeletons, the existing contrastive learning methods are usually inefficient for the acquisition of gait emotions. In this paper, we propose a contrastive learning framework utilizing selective strong augmentation (SSA) for self-supervised gait-based emotion representation, which aims to derive effective representations from limited labeled gait data. First, we propose an SSA method for the gait emotion recognition task, which includes upper body jitter and random spatiotemporal mask. The goal of SSA is to generate more diverse and targeted positive samples and prompt the model to learn more distinctive and robust feature representations. Then, we design a complementary feature fusion network (CFFN) that facilitates the integration of cross-domain information to acquire topological structural and global adaptive features. Finally, we implement the distributional divergence minimization loss to supervise the representation learning of the generally and strongly augmented queries. Our approach is validated on the Emotion-Gait (E-Gait) and Emilya datasets and outperforms the state-of-the-art methods under different evaluation protocols.

\end{abstract}


\section{Introduction}

\begin{figure}[t]
  \centerline{\includegraphics[width=3.5in]{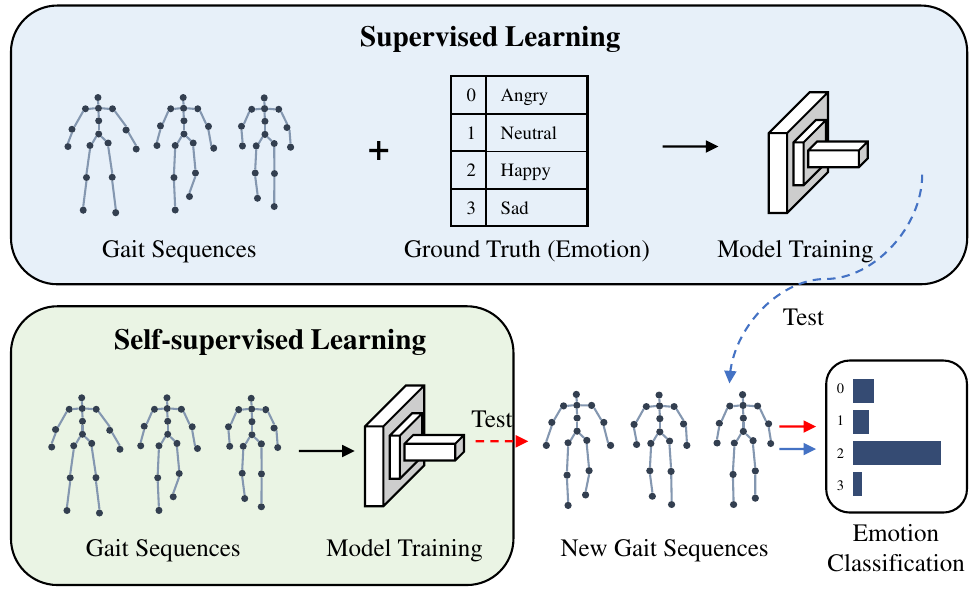}}
  \caption{The existing approaches train deep neural networks to estimate emotion classes from gait data. Supervised methods require ground-truth emotions with gait sequences for model training. Our self-supervised method trains a model from unlabeled gait sequences.}
  \label{fig1}
\end{figure}

\IEEEPARstart{E}{motions} are everywhere in the daily lives of humans and exert a significant impact on our judgment, decision-making, and behavior. Consequently, the capability for automatic emotion detection is significant in the domain of human-computer interaction\cite{cowie2001emotion} and has found extensive application in fields such as healthcare\cite{zhang2020emotion}, surveillance\cite{zhang2021real}, and robotics\cite{narayanan2020proxemo}. The existing emotion recognition research predominantly focuses on facial expressions\cite{ioannou2005emotion},\cite{xie2020facial}, \cite{umer2022facial}, speech\cite{el2011survey},\cite{chen2021novel}, text\cite{alswaidan2020survey}, \cite{acheampong2021transformer} and physiological signals such as electroencephalograms (EEGs)\cite{jenke2014feature},\cite{katsigiannis2017dreamer} and electrocardiography (ECG) signals\cite{sarkar2020self}. In cases involving the abovementioned emotion cues, facial expression-based emotion recognition methods can be unreliable when people make "mock expressions"\cite{aviezer2012body} or when self-reported emotional results are deceptive\cite{nisbett1977telling}. Furthermore, it is difficult to capture frontal facial expressions at a close range without being noticed\cite{hassan2019automatic}. Speech- and text-based methods may be less suitable for public scenes and large-scale crowds. Regarding physiological signal-based approaches, wearing specific instruments to access data is not very pragmatic.

Recently, with advancements in the fields of gait analysis and human pose estimation techniques\cite{martinez2017simple},\cite{toshev2014deeppose}, gait-based emotion recognition has attracted increasing attention. Previous studies have indicated that people experiencing different emotional states can act distinct gait behaviors\cite{kleinsmith2012affective}. Specifically, we can determine that critical gait characteristics affected by emotions involve walking velocity, step frequency, head positioning, and the extent of motion in the shoulders and elbows\cite{roether2009critical}. Compared to other emotion recognition approaches, gaits provide several unique advantages. First, considering the large range of motion during walking, gait-based methods can satisfy long-distance application scenarios and achieve nonintrusive, noncontact emotion detection. At the same time, we can acquire gait data with a webcam without overly restricting the environmental settings or requiring active cooperation from the subject. In addition, gait patterns are inherently difficult to imitate or intentionally deceive\cite{cutting1977recognizing}, making them reliable indicators for emotion recognition. Moreover, a gait-based approach involves no facial cues and requires only the positional data of the body's pivotal joints to classify emotions, which ensures people's privacy.

In gait-based emotion recognition research, earlier works focused on extracting handcrafted features. For instance, Li et al.\cite{li2016identifying} utilized the Fourier transform and statistical techniques to obtain time and frequency features for emotion classification. Bhattacharya et al.\cite{randhavane2019identifying} combined deep features extracted from a long short-term memory (LSTM) network with manually crafted affective features such as stride lengths, joint angles, and walking speeds to train a random forest classifier. As deep learning advances, an increasing number of researchers are directing their attention towards the utilization of neural network models for feature extraction and pattern recognition rather than conventional machine learning algorithms\cite{sun2022vfl}. After Yan et al.\cite{yan2018spatial} incorporated spatial-temporal graph convolutional networks (ST-GCN) into skeleton-based action recognition tasks, the effect of gait-based emotion recognition was effectively improved\cite{narayanan2020proxemo}, \cite{bhattacharya2020step}, \cite{lu2023epic}. Notably, the above approaches are supervised learning methods that rely on a substantial number of labeled data to learn emotional representations (see Fig. \ref{fig1}). However, the process of data annotation is notably labor-intensive, time-consuming, and costly, which consequently restricts the availability of labeled data. Furthermore, emotion labeling approaches are inevitably influenced by subject bias, which may lead to mislabeling. 

Contrastive learning methods that emphasize instance discrimination provide a powerful technical framework for conducting self-supervised skeleton-based representation learning. The main approach first generates positive samples through different data augmentation methods and then learns data representations by enhancing the similarity between positive samples while concurrently reducing the similarity between negative samples. Many researchers have integrated the contrastive learning paradigm into the realm of skeleton-based action recognition\cite{li20213d}, \cite{guo2022contrastive}, \cite{zhang2023hierarchical}. To identify the emotions from unlabeled gait data, Lu et al.\cite{lu2023see} first proposed a cross-coordinate contrastive learning framework named CAGE. Given an input gait sequence, they augmented it into three varying views, learned gait attributes with cross-coordinate supervision, and built a contrastive loss between the Cartesian and spherical coordinate systems. However, CAGE only applies two normal data augmentation strategies that were originally designed for action recognition tasks and does not provide any adaptive improvements for emotion recognition tasks. Designing suitable data augmentation methods is a crucial part of contrastive learning, so we must consider the characteristics of the specific task, that is, the difference between skeleton-based action recognition and emotion recognition. Moreover, the existing skeleton-based contrastive learning methods mostly adopt the ST-GCN\cite{yan2018spatial} as their encoder to process skeleton sequences\cite{shu2022multi},\cite{zeng2023contrastive}. Although the ST-GCN provides effective improvements for skeleton-based representation learning tasks, it still has some drawbacks that have not been considered by the existing research. The skeleton graph of the ST-GCN is predefined by referring to the physical structure of the human body, while the latent relationships among spatially distant joints are neglected, which can limit the representation capacities of the model. Therefore, devising an effective method that is suitable for gait-based emotion recognition and can learn representative features from unlabeled data is a significant task.

In this paper, we propose a contrastive learning framework utilizing selective strong augmentation (SSA) for self-supervised gait-based emotion representation (SSAL), which learns to optimize the encoder from multiple augmented skeleton sequences. First, we propose an SSA method that is designed specifically for the gait emotion recognition task to generate more diverse and targeted positive samples. Next, we design a complementary feature fusion network (CFFN) that integrates graph-domain and image-domain information. Finally, we implement the distributional divergence minimization loss to reduce the distributional divergence between the generally augmented samples and strongly augmented samples.

In summary, our new self-supervised learning framework for gait emotion recognition provides three key contributions.
\begin{itemize}
\item[1)] A selective strong augmentation method is proposed for the gait emotion recognition task, which incorporates upper body jitter and random spatiotemporal mask. This particular augmentation method aims to produce a more varied and focused set of positive samples, motivating the model to learn more representative and robust features.
\item[2)] A complementary feature fusion network is designed, which facilitates the integration of cross-domain information derived from the graph domain and image domain. This integration approach is intended to extract topological structural and global adaptive gait features, enhancing the generalization ability of the developed model.
\item[3)] We conduct a series of experiments on the Emotion-Gait (E-Gait)\cite{bhattacharya2020step} and Emilya datasets\cite{fourati2016perception}. The results show that our approach outperforms state-of-the-art self-supervised techniques across various evaluation protocols.
\end{itemize}

The rest of this paper is organized as follows. Section \uppercase\expandafter{\romannumeral2} reviews the previous works concerning supervised gait-based emotion recognition, self-supervised contrastive learning, and self-supervised skeleton representation. Section \uppercase\expandafter{\romannumeral3} describes the proposed method in detail. Section \uppercase\expandafter{\romannumeral4} presents the experimental details and results. Section \uppercase\expandafter{\romannumeral5} provides the conclusions of this paper.

\section{Literature Review}
\subsection{Supervised Gait-Based Emotion Recognition}

According to previous research, three general types of gait-based emotion recognition methods are available. The first type of approach utilizes sequence-based models such as recurrent neural networks (RNNs) and LSTM to learn temporal features\cite{nguyen2018skeleton},\cite{sheng2021multi}. The second category includes image-based methods that encode skeleton sequences and extract features by applying convolutional neural networks (CNNs)\cite{sun2022vfl}. In the third group, a skeleton graph is constructed in accordance with the physical structure of the human body, and a graph convolutional network (GCN) is used to explore the gait patterns of different emotions\cite{bhattacharya2020step},\cite{lu2023epic},\cite{yin2023msa}.

Among these methods, GCN-based approaches have recently received much attention because of their capacity to represent non-Euclidean data. The ST-GCN\cite{yan2018spatial}, which effectively aggregates spatiotemporal features from data, was the first model in which graph-based neural networks were applied in the domain of skeleton-based action recognition. Bhattacharya et al.\cite{bhattacharya2020step} introduced ST-GCNs to extract deep features, which were combined with manual affective features such as joint angles and velocities to perceive emotions from gaits. Sheng et al.\cite{sheng2021multi} presented a multitask learning architecture by constructing a novel attention-enhanced temporal GCN that can concurrently acquire representations for multiple objectives, such as emotion recognition, identity recognition, and auxiliary prediction. Yin et al.\cite{yin2023msa} designed skeleton data with different coarse and fine granularities and then proposed a multiscale adaptive GCN to recognize emotions. Lu et al.\cite{lu2023epic} proposed a joint reconstruction method that effectively improves the resulting classification accuracy by calculating the joint connectivity matrix based on spatiotemporal context, which exploits the latent links between body joints. The approaches mentioned above rely on supervised learning paradigms to extract affective gait features via GCNs. Considering the scarcity of available labeled emotional gait data and the possibility of mislabeling, which affects the performance and generalizability of the utilized model, we employ the self-supervised contrastive learning paradigm to learn emotional representations from unlabeled gait sequences.

\subsection{Self-Supervised Contrastive Learning}

The goal of the self-supervised learning model is to learn an effective feature embedding function from unlabeled data. Previous works\cite{zhang2016colorful},\cite{pathak2016context} concentrated on designing diverse pretext tasks to train encoders, such as rotation prediction and jigsaw puzzles. Recently, contrastive learning techniques including MoCo\cite{he2020momentum} and SimCLR\cite{chen2020simple} have shown remarkable performance compared to that of supervised learning. This type of approach applies various data augmentation strategies to generate positive samples while considering other samples as negatives relative to the input. The primary objective is to map the positive and negative sample features into a high-dimensional space and reduce the feature distances between positive pairs while increasing the feature distances between negative pairs.

In the domain of emotion recognition, contrastive learning has been accepted and utilized by many researchers. For example, Shen et al.\cite{shen2022contrastive} proposed a data-driven approach that performs contrastive learning for intersubject alignment (CLISA). The approach minimized variability across subjects by maximizing the similarity in EEG signal representations among different subjects when they received the same emotional stimuli. Mai et al.\cite{mai2022hybrid} proposed the HyCon framework for conducting hybrid contrastive learning on trimodal representations to explore interclass and intersample relationships and obtain more discriminative joint embeddings. Shuvendu Roy et al.\cite{roy2021self} introduced a contrastive learning method for multiview facial expressions (CL-MEx) to exploit facial images captured concurrently from various perspectives. Wang et al.\cite{wang2022self} presented a self-fusion contrastive learning framework, which aimed at recognizing group emotions through exploiting information acquired from faces, scenes, and objects in images. The abovementioned methods have established a strong theoretical basis for SSAL.

\subsection{Self-Supervised Skeleton Representation}

Self-supervised learning based on 3D human skeleton sequences was first applied in action recognition tasks. Rao et al.\cite{rao2021augmented} proposed a contrastive learning framework based on a momentum encoder and designed a series of novel skeleton data augmentation strategies, which laid the groundwork for subsequent research. Li et al.\cite{li20213d} explored the application of cross-view consistent knowledge as complementary supervision information to enhance the accuracy of action classification. Guo et al.\cite{guo2022contrastive} acquired abundant information from extremely augmented positive samples and forced the encoder to learn more robust action representations. Zhang et al.\cite{zhang2023hierarchical} introduced a growing data augmentation strategy along with asymmetric hierarchical learning to enhance the model performance.

For gait-based emotion recognition, Lu et al.\cite{lu2023see} first explored self-supervised learning and proposed a cross-coordinate contrastive learning framework called CAGE by constructing ambiguity samples. However, CAGE only selected two normal data augmentation methods that originated from the action recognition task. Undoubtedly, there is a discrepancy between gait-based emotion representations and skeleton-based action representations. Therefore, we must design selective augmentations that are suitable and reasonable for gait patterns. Furthermore, most skeleton-based contrastive learning methods use the ST-GCN as their encoder and focus on deep features in the graph domain while ignoring the possibility of cross-domain information fusion. Overall, we propose a contrastive learning framework utilizing selective strong augmented samples and applying a complementary feature fusion network, which can effectively learn affective representations from gait sequences.

\begin{figure*}[!t]
  \centerline{\includegraphics[width=6.7in]{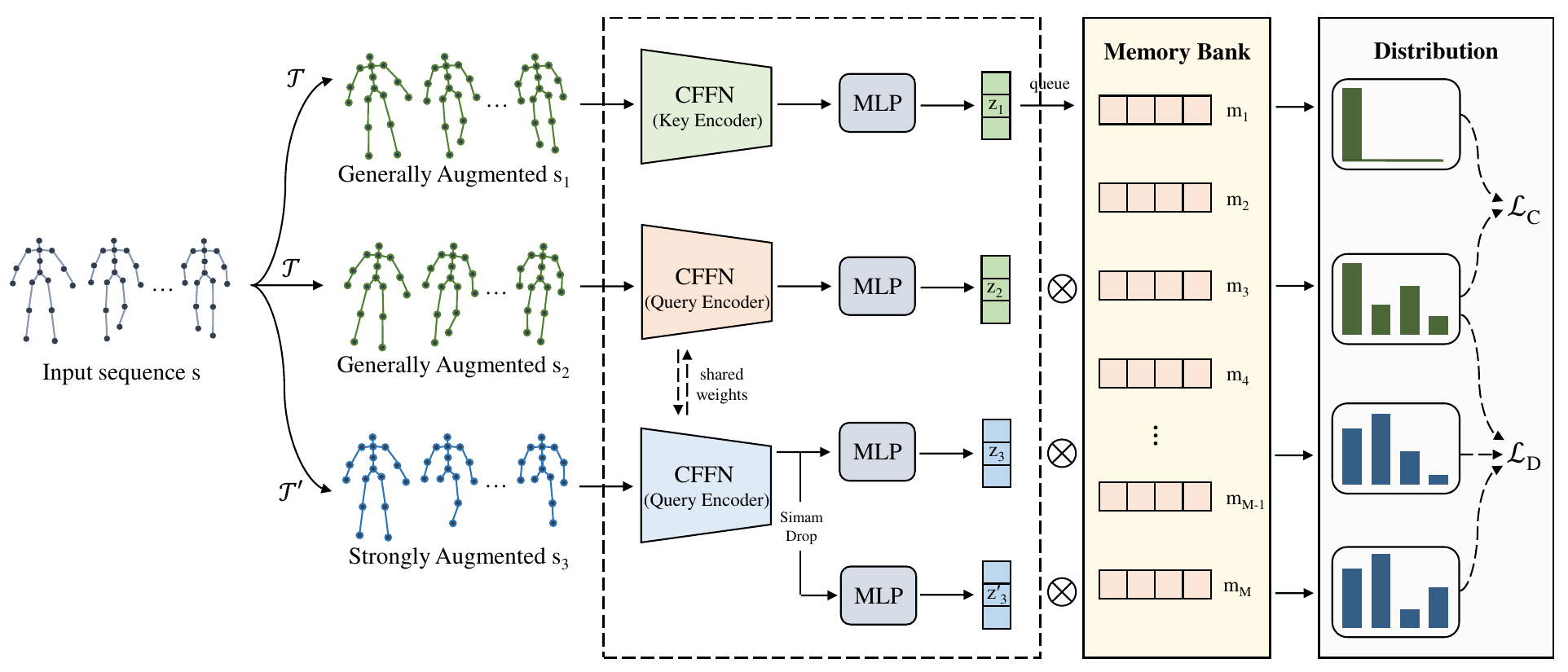}}
  \caption{The overall framework of the proposed SSAL. Given an input sequence $s$, through a \textit{general augmentation} $T$ and a \textit{strong augmentation} $T^{\prime}$, we obtain general augmentations $s_1$ and $s_2$ and a strong augmentation $s_3$. A momentum-updated key encoder and an MLP extract $z_1$, which is stored in the memory bank and serves as one of the negative samples for the subsequent training steps. The query encoder and an MLP are used to obtain $z_2$ and $z_3$, and the Simam drop is adopted to obtain ${z^{\prime}}_3$.}
  \label{fig2}
\end{figure*}

\section{Proposed Method}

\subsection{Overview}
As shown in Fig. \ref{fig2}, we propose a contrastive learning framework utilizing SSA for self-supervised gait-based emotion representation. The architecture is based on the recent advanced practice SkeletonCLR\cite{li20213d}. It applies SSA to generate positive pairs, working together with a CFFN to capture cross-domain information. Given a 3D skeleton sequence $s\in\mathbb{R}^{T\times J\times C}$ that contains $T$ consecutive frames, $J$ different body joints, and $C$ dimensions for each node, we use a general augmentation and a strong augmentation to generate positive samples $s_1$, $s_2$ and $s_3$, respectively. We feed $s_1$ into the key encoder $f_{\theta_{k}}$ and obtain a representation $\textit{f}_1$. Then, we apply a multilayer perceptron (MLP) projector $\textit{q}$ to project the representation into a lower dimension and obtain the representation $z_1$. Similarly, we feed $s_2$ into the query encoder branch to obtain the representation $z_2$. Notably, we adopt the parameter-free Simam attention module\cite{chen2021exploring} to force the model to drop several important features and learn more robust representations. Specifically, we feed $s_3$ into the query encoder, apply the drop module to the fusion features $\textit{f}_3$ and obtain the normal representation $z_3$ and the dropped representation ${z^{\prime}}_3$. A first-in-first-out dynamic memory bank $M=\{m_{i}\}_{i=1}^{M}$ is used to store the feature embeddings $z_1$, which provides negative samples for the subsequent training steps. Gradient backpropagation is employed to update the query encoder, and a moving average of the query encoder is used to update the key encoder: $\theta_{k}\leftarrow m\theta_{k}+(1-m)\theta_{q}$, where $m\in[0,1)$ is the momentum coefficient. The loss function is described in detail later.

\subsection{Selective Strong Augmentation for Skeleton}

Data augmentation is a critical approach for obtaining more positive samples, which enables the encoder to acquire abundant representations. To explore the “pattern invariance” property of skeleton sequences, we first introduce a \textit{general augmentation} strategy following previous work\cite{rao2021augmented}. It includes 3 spatial augmentations, \textit{shearing}, \textit{spatial flipping}, and \textit{rotation}, and 2 temporal augmentations, \textit{cropping} and \textit{temporal flipping}.

(1) \textit{Shearing}. To obtain positive samples with different viewpoints while retaining the original pose, we apply 3D shearing to the given skeleton sequence. The
transformation is defined as:

\begin{equation}
   \label{eq1}
S_{shearing}=X\cdot\begin{bmatrix}1&r_{12}&r_{13}\\r_{21}&1&r_{23}\\r_{31}&r_{32}&1\end{bmatrix}
\end{equation}

\noindent where \textit{X} is the original skeleton sequence, and $\{r_{12},\cdots,r_{32}\}$ are the shear factors randomly sampled from $[-1,1]$.

(2) \textit{Spatial Flipping}. Given that human gait is generally a symmetrical motion, we interchange the left and right sides of the skeleton with a probability of 0.5 to capture behavioral details.

(3) \textit{Rotation}. We apply random rotation perturbations to make the model more robust to various spatial perspectives. Specifically, we randomly choose an axis $A\in\{X, Y, Z\}$ as the principal axis and randomly rotate it by 0-30 degrees. For the remaining two axes, the rotation angles are randomly set between 0-10 degrees.

(4) \textit{Cropping}. Cropping is a temporal augmentation method that pads $T/\gamma $ frames to the original sequence and then randomly selects continuous $T$ frames to form a new sequence. $\gamma$ is the padding ratio (we set $\gamma$=2).

(5) \textit{Temporal Flipping}. Gaits are periodic, so even if we disrupt the sequence of gait, it will not affect the perception of emotions. Accordingly, we reverse the original sequence with a probability of 0.5.

\begin{figure}[t]
  \centerline{\includegraphics[width=3.5in]{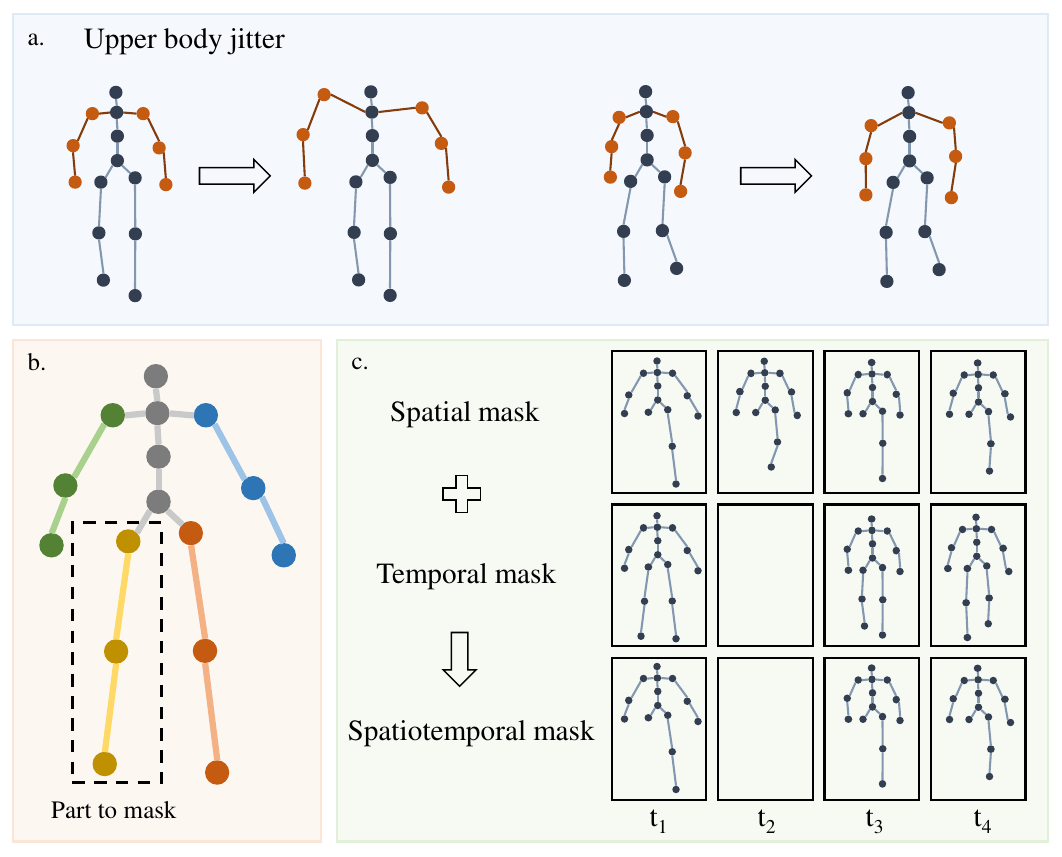}}
  \caption{Visualization of the strong augmentation. (a) We move the joints of the upper limbs to irregular positions while keeping the other joints unchanged. (b) We segment the body into five distinct parts, with each part denoted by a unique color, and then randomly mask one or two parts with zeros. (c) We apply a spatial mask to the skeleton and randomly remove several frames from the sequences, which is equivalent to a spatiotemporal mask.}
  \label{fig3}
\end{figure}

In addition to the general augmentations available for skeleton-based pattern recognition, we propose the following strong augmentations to introduce innovative and targeted patterns for emotional representation learning. Fig. \ref{fig3} shows the visualization process.

(1) \textit{Upper Body Jitter}. Previous research concluded that the movement of the upper body, especially the arms and the head, was a significant indicator of gait-based emotion recognition\cite{karg2010recognition}. Therefore, we consider applying an upper body jitter to transform the joint positions to motivate the model to learn representative features. Specifically, we select the upper body joints (shoulders, elbows, and hands) and move these joints to irregular positions while keeping the other joints unchanged. The transformation is defined as follows: 

\begin{equation}
   \label{eq2}
S_{jitter}=X[:,j]\cdot\begin{bmatrix}r_{11}&r_{12}&r_{13}\\r_{21}&r_{22}&r_{23}\\r_{31}&r_{32}&r_{33}\end{bmatrix}
\end{equation}

\noindent where $j$ is the upper body joint set, and $\{r_{11},\cdots,r_{33}\}$ are the jitter parameters randomly sampled from $[-1,1]$.

(2) \textit{Random Spatiotemporal Mask}. Inspired by the observation that we can also recognize emotion from incomplete gait sequences that lack some time frames and body parts, we propose a random spatiotemporal mask to make the model learn more robust feature representations. 

To generate a spatial mask, we first divide the human skeleton into five body components, the limbs and the torso, which can efficiently reflect body movements. Then, we randomly select one or two of these parts and replace the coordinates of the joints with zeros. The spatial mask formula is as follows:

\begin{equation}
   \label{eq3}
S_{spatial}(X)=X\astrosun Mask_s(RanSamp(part))
\end{equation}

\noindent where $S_{spatial}(X)$ is the skeleton joint matrix after applying spatial mask augmentation. $X$ is the input skeleton joint matrix. $\astrosun$ is the dot product operation. $RanSamp(\cdot)$ is the random sampling function that randomly selects one or two parts from the predefined sets. $Mask_s(\cdot)$ is the spatial mask function that transforms the joint coordinates of the selected part set to zero.

The temporal mask is the same. We randomly select several frames and mask all the joints with zeros. Therefore, the spatiotemporal mask formula is:

\begin{equation}
   \label{eq4}
S_{st}(X)=S_{spatial}(X)\astrosun Mask_t\big(RanSamp(r\times T)\big)
\end{equation}

\noindent where $S_{st}(X)$ is the skeleton joint matrix obtained after applying the spatiotemporal mask augmentation. $RanSamp(\cdot)$ is a random sampling function that randomly selects several frames from the original frames. $Mask_t(\cdot)$ is the temporal mask function that transforms the joint coordinates of the selected frames to zeros. $r$ is the temporal mask parameter (we set $r$=0.25), and $T$ is the number of frames.

\subsection{Complementary Feature Fusion Network}
Most of the previously developed skeleton-based contrastive learning frameworks adopt the ST-GCN as their encoder to learn representations. We can see that GCNs have great advantages in terms of processing non-Euclidean data such as human skeleton sequences, but some problems remain. First, the degree of freedom of the human body is so complicated that applying the same adjacency matrices to the channels would limit the ability of the model to address the dependency correlations of joints. Second, the spatial-temporal graph connects only the same joints in different frames, thus the latent links among distant joints in successive frames are neglected. To focus on the global features in the spatial and temporal dimensions, we propose a CFFN, which integrates the cross-domain information derived from the graph domain and image domain to learn complementary feature representations. Specifically, we adopt the ST-GCN as the graph-domain feature extractor to obtain topological structural information. Moreover, we introduce an adaptive frequency filter (AFF)-based token mixer\cite{huang2023adaptive} as the image-domain feature extractor to obtain global adaptive representations. The AFF token mixer utilizes a Fourier transform to transfer a latent representation to the frequency domain and employs elementwise multiplication to realize semantic-adaptive frequency filtering.

\begin{figure}[t]
  \centerline{\includegraphics[width=3.5in]{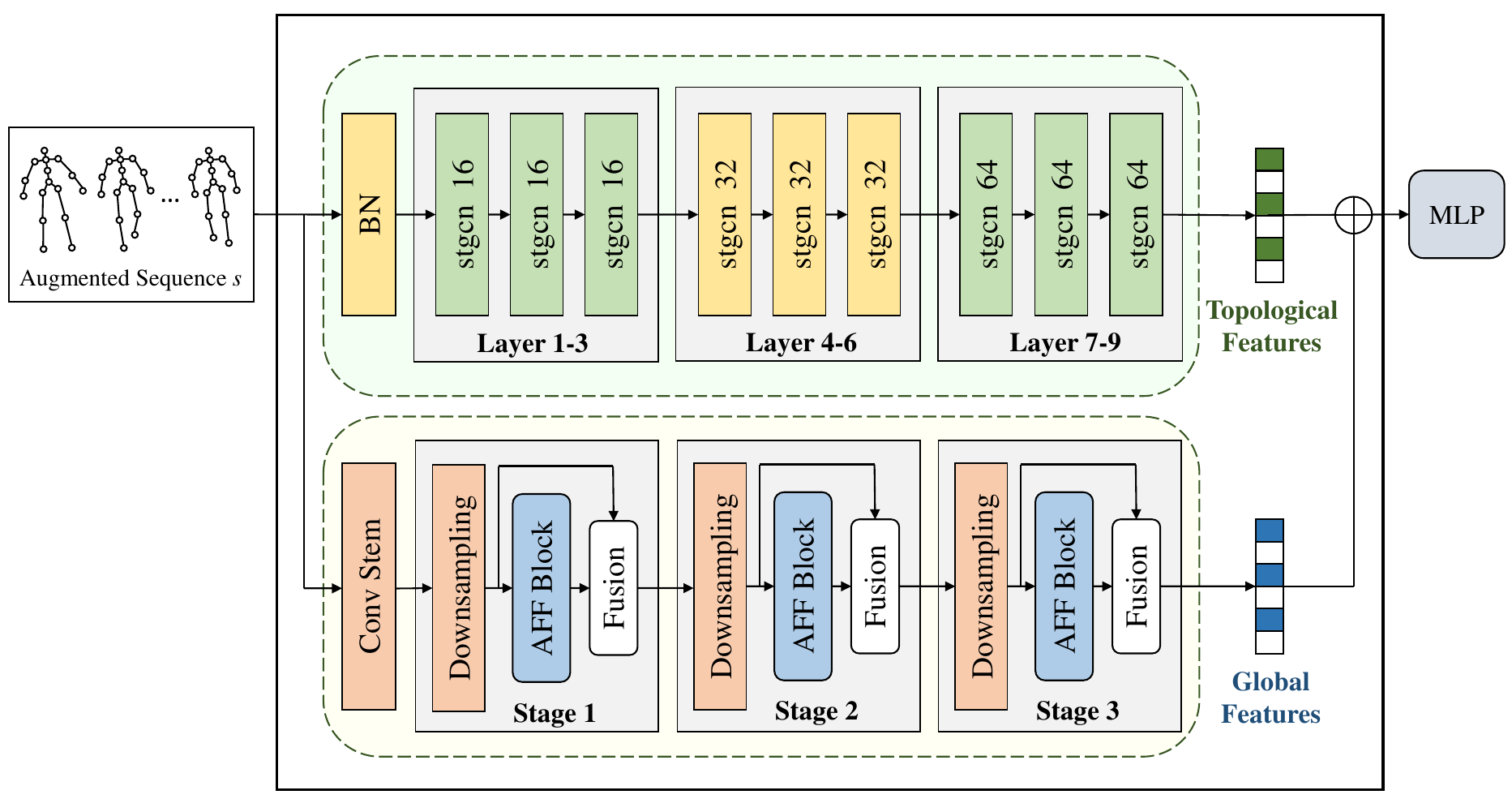}}
  \caption{The architecture of the proposed CFFN. The graph-domain branch is designed with reference to the ST-GCN. The image-domain branch applies an AFF token mixer. Finally, we obtain a 128-dimensional fusion feature vector.}
  \label{fig4}
\end{figure}

The architecture of the CFFN is shown in Fig. \ref{fig4}. For the graph-domain branch, we first apply batch normalization to ensure that the scale of the input augmented sequences is consistent across different joints. The backbone is composed of 9 layers of spatial-temporal graph convolution operators (ST-GCN units). The initial 3 layers, the subsequent 3 layers, and the last 3 layers have 16, 32, and 64 output channels, respectively. The temporal kernel size is 9, and the spatial kernel size is 3. The strides of the 4th and 7th temporal convolution layers are 2, and the strides of the other layers are 1. The final dimensionality of the topological structural features is 64.

For the image-domain branch, we first employ a convolution stem for tokenization purposes. At each stage, we apply layer normalization (LN) to the input for channel mixing and then feed the result to the AFF token mixer for global token mixing to obtain the output of the AFF block. Then, we use plain fusion to connect the local and global features. The backbone network of AFFNet is composed of multiple AFF blocks. The final dimensionality of the global adaptive features is 64.

We concatenate these two features directly and obtain a 128-dimensional feature vector. Then, we employ a two-layer nonlinear MLP to project the integrated features to a lower-dimensional space.

\subsection{Loss Function}

The purpose of SSAL is to learn effective emotional representations by contrasting multiple gait sequences. The model is expected to amplify the similarity between the original sequence and the augmented sequences while reducing the similarity between the positive and negative samples in the memory bank. In prior works, the contrastive InfoNCE loss was defined as:

\begin{equation}
   \label{eq5}
  {\mathcal L}_{c}=-log\frac{exp(z\cdot z\prime/\tau)}{exp(z\cdot z\prime/\tau)+\sum_{i=1}^{M}exp(z\cdot m_{i}/\tau)}
\end{equation}

\noindent where M is the length of the memory queue, $m_i$ is the \textit{i}-th negative sample and $\tau$ is the temperature hyperparameter. 

Considering the dramatic discrepancy between the movement patterns of the generally and strongly augmented sequences, \cite{wang2022contrastive} indicated that for a randomly initialized network, the generally augmented sequence and the strongly augmented sequence possess similar distributions. Thus, we can obtain the following conditional distributions:
\begin{equation}
   \label{eq6}
  p(z_1|z_2)=\frac{exp(z_1\cdot z_2/\tau)}{exp(z_1\cdot z_2/\tau)+\sum_{i=1}^Mexp(z_2\cdot m_i/\tau)}
\end{equation}

\begin{equation}
   \label{eq7}
  p(m_i|z_2)=\frac{exp(m_i\cdot z_2/\tau)}{exp(z_1\cdot z_2/\tau)+\sum_{i=1}^Mexp(z_2\cdot m_i/\tau)}
\end{equation}

\noindent where $p(z_1|z_2)$ and $p(m_i|z_2)$ represent the likelihood of the query representation $z_2$ being assigned to its positive counterpart $z_1$ and to the embedding $m_i$ in the memory bank M, respectively. To minimize the distributional divergence between a generally augmented sequence and a strongly augmented sequence, the loss can be written as follows:

\begin{equation}
\begin{split}
   \label{eq8}
  L_{d1}=&-p(z_{1}|z_{2})logp(z_{1}|z_{3})\\
  &-\sum_{i=1}^{M}p(m_{i}|z_{2})logp(m_{i}|z_{3})
\end{split}
\end{equation}

As mentioned earlier, we adopted the parameter-free Simam attention module. The distributional divergence between a generally augmented sample and a dropped strongly augmented sample is the same:
\begin{equation}
\begin{split}
   \label{eq9}
  L_{d2}=&-p(z_{1}|z_{2})logp(z_{1}|z_{3}^{\prime})\\
  &-\sum_{i=1}^{M}p(m_{i}|z_{2})logp(m_{i}|z_{3}^{\prime})
\end{split}
\end{equation}

Therefore, the distributional divergence loss can be given by 
\begin{equation}
   \label{eq10}
  {\mathcal L}_{d}=1/2({\mathcal L}_{d1}+{\mathcal L}_{d2})
\end{equation}

The overall loss for our SSAL method can be formulated as $\mathcal{L}=\alpha\mathcal{L}_{Info}+\beta\mathcal{L}_{d}$, where $\alpha$ and $\beta$ are the coefficients used to balance the loss. Here, we set $\alpha$=$\beta$=1 to obtain a more general model.

\section{Experiments}
In Section A, two public datasets used in the experiments are described. In Section B, the experimental settings of SSAL are presented. In Section C, the evaluation criteria are declared. In Section D, the comparison results with state-of-the-art methods are displayed. In Section E, the ablation results on each part are discussed.

\subsection{Datasets}
1) E-Gait\cite{bhattacharya2020step} includes 2,177 real gaits, and each gait is labeled with one of the four emotion classes (angry, neutral, happy, or sad) by the same 10 annotators. Specifically, the dataset is composed of two parts. Part 1 contains 342 gaits collected from diverse sources, including BML\cite{ma2006motion}, Human3.6M\cite{ionescu2013human3}, ICT\cite{narang2017motion}, and CMU-MOCAP\cite{sigal2010humaneva}. Part 2 is derived from ELMD\cite{komura2017recurrent} and consists of 1,835 real gait sequences.

2) Emilya\cite{fourati2016perception} is a dataset of emotional body expressions concerning different daily actions. It contains 7 daily actions, including simple walking (SW), walking with an object in hands (WH), sitting down (SD), knocking at the door (KD), moving books on a table with two hands (MB), lifting an object (Lf) and throwing an object (Th). Twelve actors were asked to perform the actions with 8 emotions, including anxiety (AX), pride (Pr), joy (Jy), sadness (Sd), panic/fear (PF), shame (Sh), anger (Ag) and neutral (Nt). We select the motion capture data of simple walking with 4 emotions (anger, neutral, joy, and sadness).

We uniformly convert the skeleton data into 16 body joints and 120 frames. For the E-Gait dataset, we randomly split the training and testing sets at a ratio of 4:1. As for the Emilya dataset, the data of 9 actors are allocated for training, and the remaining is used for testing. To determine the distribution of the data, we calculate the percentage of each emotional class contained in the dataset. As shown in Table~\ref{table1}, the E-Gait dataset contains a few gait data with sad labels, and angry gaits account for more than half of the dataset. The Emilya dataset is relatively balanced across all emotion labels.

\begin{table}[h]
    \caption{The distribution of each kind of emotion}
    \renewcommand\arraystretch{1.2}
    \label{table1}
    \begin{center}
    \setlength{\tabcolsep}{3.2mm}{
        \begin{tabular}{c c c c c}
            \toprule
            {\bf Dataset} &{\bf Angry} &{\bf Neutral} &{\bf Happy} &{\bf Sad}\\      
            \midrule  	
            E-Gait & 55.03$\%$ & 23.45$\%$ & 14.61$\%$ & 6.90$\%$ \\
            Emilya & 19.63$\%$ & 21.18$\%$ & 22.80$\%$ & 36.38$\%$ \\
            \bottomrule
        \end{tabular}}
    \end{center}
\end{table}

\subsection{Experimental Settings}

We adopt the PyTorch framework to implement the proposed method and conduct all the experiments on an Ubuntu server equipped with an Intel Xeon@2.16 GHz CPU and 4 NVIDIA GTX Titan X graphics cards.

\textbf{Data Augmentation.} We compare different general augmentation strategy compositions and select the two most effective general augmentations. By applying general augmentation and SSA to the input skeleton data, we explore the effect of SSA.

\textbf{Self-supervised Pretext Training.} For the contrastive learning parameter settings, we follow those used in AimCLR\cite{guo2022contrastive}. In particular, the feature dimensionality is 128, the size of the memory bank M is 2560, the momentum coefficient m is 0.999, and the temperature hyperparameter $\tau$ is 0.07. For optimization, we employ stochastic gradient descent (SGD) with a momentum of 0.9 and a weight decay of 0.0001.
We adopt the CFFN  as the encoder. The model is trained for 500 epochs with an initial learning rate of 0.001 (which is multiplied by 0.1 at epoch 400).

\textbf{Linear Evaluation Protocol.} To verify the effectiveness of the representations learned from the pretext training for the gait-based emotion recognition task, we train a linear classifier on labeled datasets. Specifically, we freeze the encoder parameters and train a linear classifier, which consists of a fully connected layer and a softmax layer. The classifier is trained for 200 epochs with an initial learning rate of 0.001 (which is multiplied by 0.1 at epoch 100).

\textbf{Finetuned Evaluation Protocol.} We append a linear classifier to the trained encoder and train the entire model in a supervised training mode to optimize the performance of the model. The model is trained for 100 epochs with an initial learning rate of 0.0001 (which is multiplied by 0.1 at epoch 50).

\textbf{Semi-supervised Evaluation Protocol.} We fine-tune the pre-trained encoder with only 5$\%$, 10$\%$, 20$\%$, and 50$\%$ of the labeled data, and the employed data are randomly selected. The model is trained for 20 epochs with an initial learning rate of 0.001 (which is multiplied by 0.1 at epoch 10).

\subsection{Evaluation Criteria}
To evaluate the performance of the proposed SSAL algorithm in the gait-based emotion classification tasks, we calculate the classification accuracy, precision, recall, and F1 score via the following formulas:

\begin{equation}
   \label{eq11}
Accuracy=\frac{{TP}+{TN}}{{TD}}
\end{equation}

\begin{equation}
   \label{eq12}
Precision=\sum(\frac{TP_{i}}{TP_{i}+FP_{i}}*w_{i})
\end{equation}

\begin{equation}
   \label{eq13}
Recall=\sum(\frac{TP_{i}}{TP_{i}+FN_{i}}*w_{i})
\end{equation}

\begin{equation}
   \label{eq14}
F1 score=\sum\frac{2*Precision_i*Recall_i}{(Precision_i+Recall_i)}
\end{equation}

\noindent where TP, FP, TN, and FN represent the numbers of true positives, false positives, true negatives, and false negatives for the four emotions, respectively. TD represents the total number of data. $w_{i}$ represents the number of samples in each class as a proportion of the total number of samples in all classes, and \textit{i} = 0, 1, 2, 3.

\begin{table}[t]
    \caption{Linear evaluation results on the E-Gait dataset}
    \renewcommand\arraystretch{1.2}
    \label{table2}
    \begin{center}
    \setlength{\tabcolsep}{2mm}{
        \begin{tabular}{l| c| c| c| c}
            \toprule
            {\bf Method} & {\bf Accuracy} & {\bf Precision} & {\bf Recall} & {\bf F1} \\
            \midrule
            \textit{Supervised}        & & & & \\
            ST-GCN\cite{yan2018spatial}         & 75.47 & 78.22 & 75.47 & 75.22  \\
            STEP\cite{bhattacharya2020step}     & 80.95 & 81.06 & 80.20 & 79.81  \\  \hline
            \textit{Self-supervised}      & & & & \\      
            CrosSCLR\cite{li20213d}          & 79.33 & 78.60 & 79.33 & 78.63  \\
            AimCLR\cite{guo2022contrastive}     & 78.95 & 77.57 & 78.95 & 78.00  \\
            HiCLR\cite{zhang2023hierarchical}   & 80.32 & 80.96 & 80.32 & 79.99  \\
            CAGE\cite{lu2023see}                & 79.59 & --    & --    & --  \\
            \textbf{SSAL(Ours)}       & \textbf{81.12} & \textbf{81.89} & \textbf{81.25} & \textbf{80.72} \\
            \bottomrule
        \end{tabular}}
    \end{center}
\end{table}

\begin{table}[ht]
    \caption{Linear evaluation results on the Emilya dataset}
    \renewcommand\arraystretch{1.2}
    \label{table3}
    \begin{center}
    \setlength{\tabcolsep}{2mm}{
        \begin{tabular}{l| c| c| c| c}
            \toprule
            {\bf Method} & {\bf Accuracy} & {\bf Precision} & {\bf Recall} & {\bf F1} \\
            \midrule
            \textit{Supervised}        & & & & \\      
            ST-GCN\cite{yan2018spatial}         & 65.98 & 69.31 & 65.98 & 67.03  \\
            STEP\cite{bhattacharya2020step}     & 70.77 & 65.08 & 54.36 & 52.30  \\  \hline
            \textit{Self-supervised}      & & & & \\      
            CrosSCLR\cite{li20213d}          & 66.50 & 63.04 & 66.50 & 60.00  \\
            AimCLR\cite{guo2022contrastive}     & 60.36 & 63.91 & 60.36 & 59.81  \\
            HiCLR\cite{zhang2023hierarchical}   & 67.77 & 71.72 & 67.77 & 68.24  \\
            \textbf{SSAL(Ours)}       & \textbf{76.04} & \textbf{75.20} & \textbf{75.78} & \textbf{74.49} \\
            \bottomrule
        \end{tabular}}
    \end{center}
\end{table}

\subsection{Comparison with State-of-the-art}
Since few self-supervised methods are available for gait-based emotion recognition, we compare the proposed SSAL with related skeleton-based contrastive learning methods that operate similarly in gait-based emotion recognition tasks.

\textbf{Linear Evaluation Results on the E-Gait Dataset.}  We conduct an extensive comparison with previously developed supervised methods and recent methods for skeleton-based self-supervised action recognition. As shown in Table~\ref{table2}, the accuracy of SSAL is improved by 0.80$\%$-2.17$\%$ over those of the existing contrastive learning methods. Even compared to some supervised methods, our approach achieves superior performance. Moreover, the SSAL achieves the best results in terms of precision, recall, and F1 score, indicating that our method has a high classification learning capacity.

\textbf{Linear Evaluation Results on the Emilya Dataset.} Table~\ref{table3} shows that our proposed SSAL outperforms all other self-supervised methods and supervised methods in terms of accuracy, precision, recall, and F1 score. Specifically, compared to the advanced contrastive learning methods, AimCLR\cite{guo2022contrastive} and HiCLR\cite{zhang2023hierarchical}, our approach provides accuracy improvements of 15.68$\%$ and 8.27$\%$, respectively. Notably, the Emilya dataset has approximately half the data size of the E-Gait dataset. The results show that for the gait-based emotion recognition task, SSAL has great advantages in small sample datasets over the existing skeleton-based methods.

\textbf{Finetuned Evaluation Results.} We compare the finetuned evaluation results on the E-Gait and Emilya datasets. The experimental setup is consistent with CAGE, and the model trains 20 epochs. Table~\ref{table4} shows that our proposed SSAL achieves the best performance. Specifically, on the E-gait dataset, SSAL surpasses the self-supervised gait emotion recognition method CAGE by 0.05$\%$. Compared with the latest contrastive learning method HiCLR, the accuracy of SSAL on the E-Gait and Emilya datasets is improved by 0.30$\%$ and 6.75$\%$, respectively.

\textbf{Semi-supervised Evaluation Results.} We use a small amount of labeled data for semi-supervised evaluation and compare our approach with other outstanding methods on the E-Gait and Emilya datasets. Table~\ref{table5} shows that in all cases, our proposed SSAL approach outperforms the other advanced methods. In particular, with only 5$\%$ annotated data, SSAL achieves accuracies of 76.75$\%$ and 64.58$\%$ on the E-Gait and Emilya datasets, respectively, indicating that our approach has a significant advantage in terms of learning from only a small quantity of labeled data.

\begin{table*}[ht]
\caption{Semi-supervised evaluation results on the E-Gait and Emilya datasets}
\label{table5}
\setlength{\tabcolsep}{6mm}{
\begin{center}
\begin{tabular}{l|cccc|cccc}
\toprule
\multirow{2}{*}{Method} & \multicolumn{4}{c|}{E-Gait($\%$)}  & \multicolumn{4}{c}{Emilya($\%$)}\\
\cmidrule(l){2-9}  
& 5$\%$   & 10$\%$  & 20$\%$   & 50$\%$  & 5$\%$  & 10$\%$   & 20$\%$  & 50$\%$ \\
\midrule
CrosSCLR\cite{li20213d}        & 62.94 & 72.89 & 75.62 & 78.86 & 57.89 & 60.53 & 62.82 & 65.13 \\
AimCLR\cite{guo2022contrastive}   & 71.48 & 75.72 & 76.84 & 78.46 & 47.37 & 52.63 & 55.13 & 60.00 \\
HiCLR\cite{zhang2023hierarchical} & 69.61 & 74.35 & 78.83 & 79.58 & 57.80 & 60.87 & 62.92 & 65.73 \\
CAGE\cite{lu2023see}              & 70.64 & 78.90 & 79.13 & 81.65 & --    & --    & --    & --    \\
\textbf{SSAL(Ours)} & \textbf{76.75} & \textbf{79.75} & \textbf{80.25} & \textbf{82.00}  & \textbf{64.58}  & \textbf{68.23} & \textbf{70.83} & \textbf{72.40} \\
\bottomrule
\end{tabular}
\end{center}}
\end{table*}

\begin{table}[t]
    \caption{Finetuned evaluation results on the E-Gait and Emilya datasets}
    \renewcommand\arraystretch{1.2}
    \label{table4}
    \begin{center}
    \setlength{\tabcolsep}{4mm}{
        \begin{tabular}{l| c| c}
            \toprule
            {\bf Method} & {\bf E-Gait ($\%$)} & {\bf Emilya ($\%$)} \\
            \midrule
            CrosSCLR\cite{li20213d}          &          81.82    &         69.31 \\
            AimCLR\cite{guo2022contrastive}     &          81.20    &         63.68 \\
            HiCLR\cite{zhang2023hierarchical}   &          82.32    &         70.59 \\
            CAGE\cite{lu2023see}                &          82.57    &         --    \\ 
            \textbf{SSAL(Ours)} &  \textbf{82.62} &   \textbf{77.34} \\
            \bottomrule
        \end{tabular}}
    \end{center}
\end{table}

\begin{figure}[t]
  \centerline{\includegraphics[width=3in]{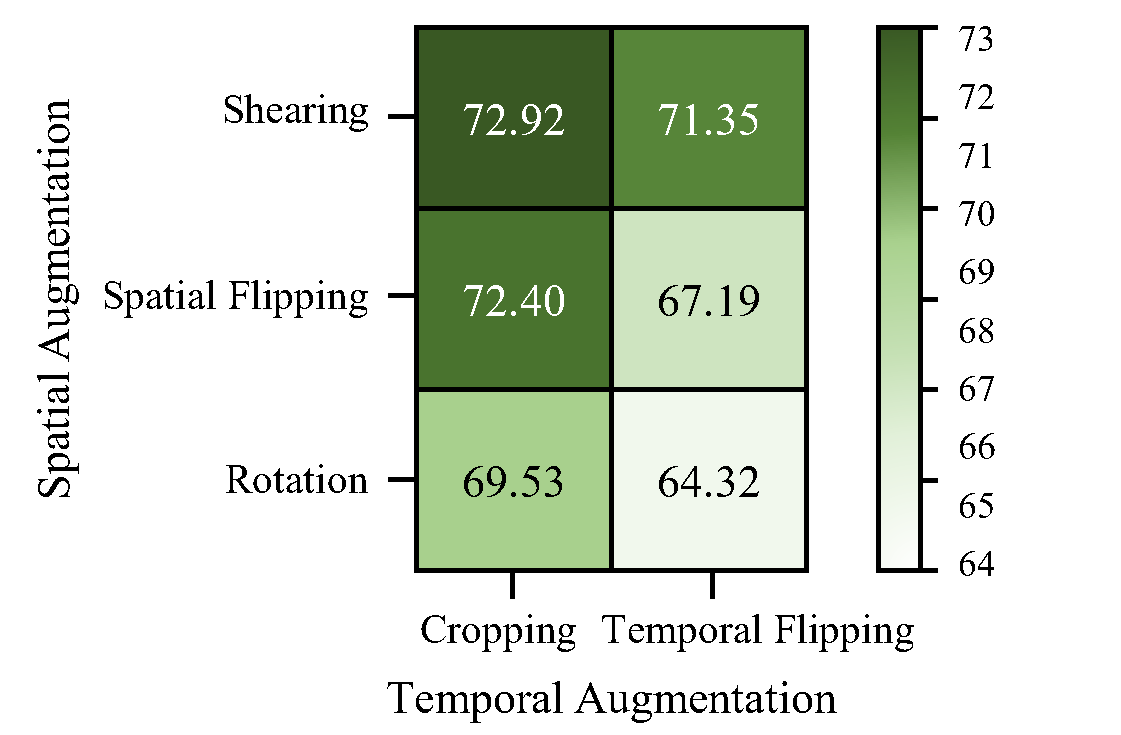}}
  \caption{Top-1 accuracy achieved with different general augmentation strategy compositions on the Emilya dataset.}
  \label{fig5}
\end{figure}

\begin{table}[t]
    \caption{Ablation study results concerning the data augmentation method}
    \renewcommand\arraystretch{1.2}
    \label{table6}
    \begin{center}
    \setlength{\tabcolsep}{3.2mm}{
        \begin{tabular}{c c c c c}
            \toprule
            {\bf GA} & {\bf UBJ} & {\bf RSM} &  {\bf E-Gait ($\%$)} &  {\bf Emilya ($\%$)} \\
            \midrule
             \checkmark &          - &           - &          78.25 &      72.92 \\
             \checkmark &      \checkmark &      - &          81.00 &      73.44 \\
             \checkmark &          - &       \checkmark &      80.88 &      75.00 \\
             \textbf{\checkmark} &      \textbf{\checkmark} &  \textbf{\checkmark} &       \textbf{81.12} &      \textbf{76.04} \\
            \bottomrule
        \end{tabular}}
    \end{center}
\end{table}

\begin{table}[t]
    \caption{Ablation study results concerning the encoder network}
    \renewcommand\arraystretch{1.2}
    \label{table7}
    \begin{center}
    \setlength{\tabcolsep}{3.2mm}{
        \begin{tabular}{c c c}
            \toprule
            {\bf Method} & {\bf E-Gait ($\%$)} & {\bf Emilya ($\%$)} \\
            \midrule
              graph domain &                80.38 &              74.74 \\
              image domain &                80.75 &              67.71 \\
            \textbf{CFFN} &                 \textbf{81.12} &   \textbf{76.04} \\
            \bottomrule
        \end{tabular}}
    \end{center}
\end{table}

\begin{figure}[t]
  \centerline{\includegraphics[width=3.5in]{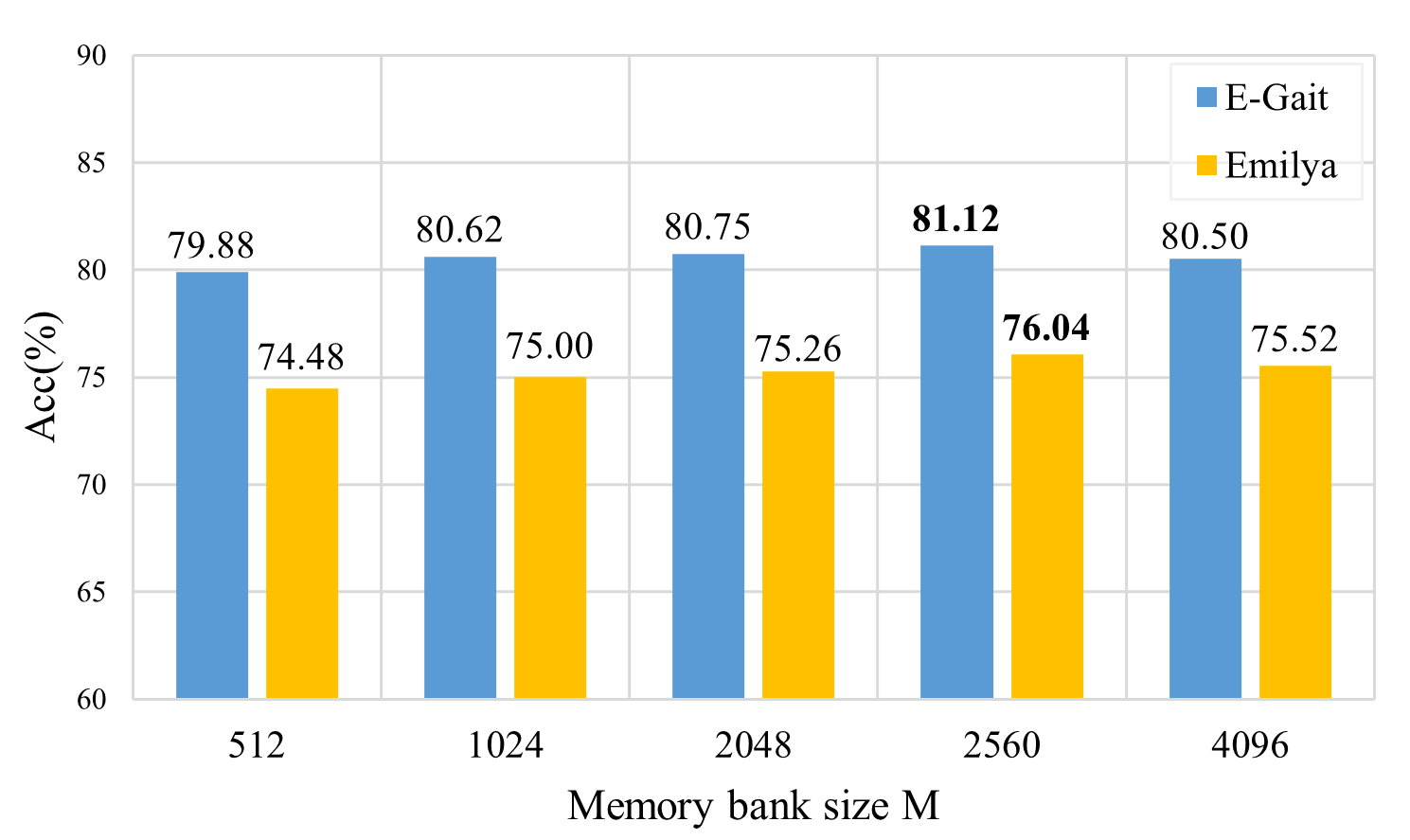}}
  \caption{Comparison among the top-1 accuracy achieved with different memory bank sizes M on the E-Gait and Emilya datasets.}
  \label{fig6}
\end{figure}

\subsection{Ablation Study}

We conduct ablation experiments to validate the efficiency of the different components of our method. All the experiments follow the self-supervised pretext training and linear evaluation protocol.

\textbf{The effectiveness of SSA.} We first take the Emilya dataset as an example to select the two most effective general augmentations among the five methods described. As shown in Fig. \ref{fig5}, we combine each of the three spatial augmentations and the two temporal augmentations individually. Of the six compositional strategies, the combination of "Shearing" and "Cropping" performs best, and this is consistent with the data augmentation methods used in previous experiments\cite{li20213d},\cite{guo2022contrastive}.

On this basis, we compare the effects of introducing SSA and other augmentations. As shown in Table~\ref{table6}, after applying upper body jitter (UBJ), the accuracies are improved by 2.75$\%$ and 0.52$\%$ on the E-Gait and Emilya datasets, respectively, demonstrating that the arms and the head are significant emotional clues. Notably, the random spatiotemporal mask (RSM) performs better, which shows that the models learn high-level semantic information in the spatial and temporal dimensions. When the UBJ and the RSM are used, our proposed SSAL approach achieves the best results.

\textbf{The effectiveness of the CFFN.} We explore the effectiveness of the graph domain, image domain, and CFFN. As shown in Table~\ref{table7}, our proposed CFFN integrates cross-domain information and reaches the highest accuracy levels on the E-Gait and Emilya datasets. Especially on the Emilya dataset, the CFFN improves the final accuracies by 1.30$\%$ and 8.33$\%$, respectively. This shows that the CFFN has a great capacity to aggregate representative features in the spatial and temporal dimensions and provide global adaptive information about the target skeleton, helping the encoder learn more robust and representative features for downstream tasks.

\textbf{The effectiveness of different memory bank sizes.} As shown in Fig. \ref{fig6}, we compare the model performances attained with different memory bank sizes. A large memory bank yields better performance, and our proposed SSAL method obtains the best result when M = 2560. However, when the size of the memory bank reaches a certain level, the number of negatives becomes much larger than that of positives, which may lead to a shortcut during representation learning.

\begin{figure}[t]
  \centerline{\includegraphics[width=3.5in]{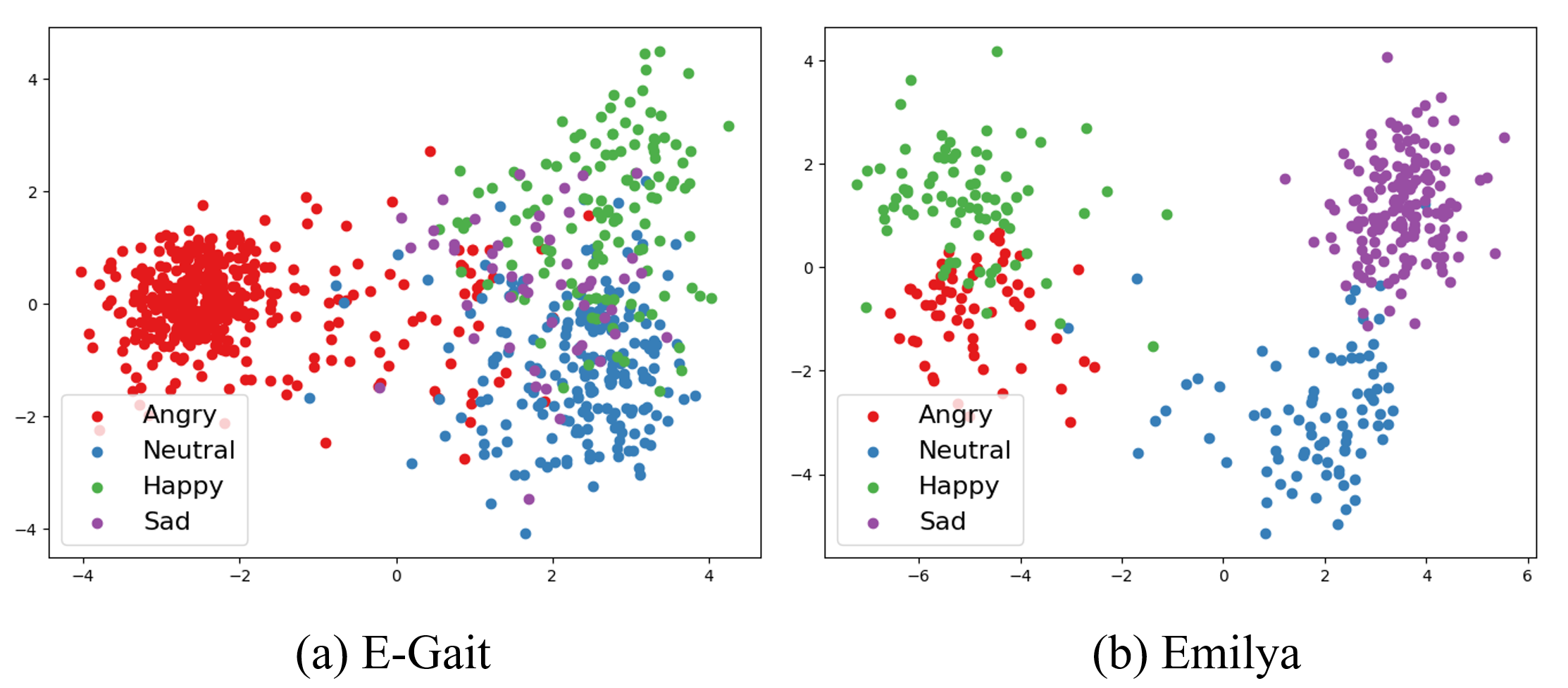}}
  \caption{(a) LDA visualization of the embeddings produced on the E-Gait dataset. (b) LDA visualization of the embeddings produced on the Emilya dataset.}
  \label{fig7}
\end{figure}

\textbf{Qualitative Results.} We apply latent Dirichlet allocation (LDA)\cite{blei2003latent} to show the embedding distributions of SSAL. The results are fair comparisons conducted over 500 epochs of pretraining on the E-Gait and Emilya datasets. In Fig. \ref{fig7}, the embeddings of SSAL exhibit tight clustering across both datasets, which verifies that SSAL can generate discriminative features to recognize different emotions accurately.

\section{Conclusion}

In this paper, we propose a contrastive learning framework SSAL, which utilizes SSA to predict emotion classes from unlabeled gait data. Specifically, upper body jitter and random spatiotemporal mask are used as SSAs together with the general shearing and cropping augmentations to generate positive samples. The CFFN is proposed to extract complementary fusion features, which aggregate cross-domain topological structural and global adaptive representations. Experimental results obtained on the E-Gait and Emilya datasets demonstrate the promising performance of SSAL under a variety of evaluation protocols.

This study has several limitations. First, the amount of available labeled emotional gait data is limited, and these data are relatively unbalanced. If more data were available, the performance of the proposed model could be further improved. Second, the proposed SSAL approach considers only unimodal gait data, and we can use more information to support the process of classifying emotions in a given application scenario. In the future, it is anticipated that the aforementioned limitations will be addressed to develop a more precise and robust approach for emotion recognition tasks.


\bibliographystyle{IEEEtran}
\bibliography{IEEEabrv,paper1.bib}

\end{document}